\UseRawInputEncoding
\pdfoutput=1 
\documentclass[10pt,twocolumn,letterpaper]{article}
\usepackage{iccv}
\usepackage{times}
\usepackage{epsfig}
\usepackage{graphicx}
\usepackage{amsmath}
\usepackage{amssymb}
\usepackage[accsupp]{axessibility}  
\usepackage{booktabs}
\usepackage[mathscr]{eucal}
\usepackage{caption}
\usepackage{subcaption, multirow, makecell, cite}
\usepackage[pagebackref=true,breaklinks=true,letterpaper=true,colorlinks,bookmarks=false]{hyperref}

\usepackage[breaklinks=true,bookmarks=false]{hyperref}

\iccvfinalcopy 

\def\iccvPaperID{9553} 

\ificcvfinal\fi

\makeatletter
\def\thanks#1{\protected@xdef\@thanks{\@thanks
		\protect\footnotetext{#1}}}
\makeatother

\begin{document}

\title{RFD-ECNet: Extreme Underwater Image Compression \\with Reference to Feature Dictionary}

%
\author{
	Mengyao Li$^{1}$, Liquan Shen$^{1*}$, Peng Ye$^2$, Guorui Feng$^{1}$, Zheyin Wang$^{1}$ \\
	$^{1}$Shanghai Universit, Shanghai, China    
	$^{2}$Fudan University, Shanghai, China \\
	{\tt\small sdlmy@shu.edu.cn, jsslq@163.com, 20110720039@fudan.edu.cn} \\ {\tt\small grfeng@shu.edu.cn, wangzy$\_$0831@163.com}
	\thanks{* Corresponding Author}
}

%
%
%
%

\maketitle
\ificcvfinal\thispagestyle{empty}\fi

\begin{abstract} 
	\vspace{-2pt}  
	Thriving underwater applications demand efficient extreme compression technology to realize the transmission of underwater images (UWIs) in very narrow underwater bandwidth.
	However, existing image compression methods achieve inferior performance on UWIs because they do not consider the characteristics of UWIs:
	(1) Multifarious underwater styles of color shift and distance-dependent clarity, caused by the unique underwater physical imaging; (2) Massive redundancy between different UWIs, caused by the fact that different UWIs contain several common ocean objects, which have plenty of similarities in structures and semantics. To remove redundancy among UWIs, we first construct an exhaustive underwater multi-scale feature dictionary to provide coarse-to-fine reference features for UWI compression.
	Subsequently, an extreme UWI compression network with reference to the feature dictionary (RFD-ECNet)\footnote{Code is available at \scriptsize{\url{https://github.com/lilala0/RFD-ECNet}}} is creatively proposed, which utilizes feature match and reference feature variant to significantly remove redundancy among UWIs. 
	To align the multifarious underwater styles and improve the accuracy of feature match, an underwater style normalized block (USNB) is proposed, which utilizes underwater physical priors extracted from the underwater physical imaging model to normalize the underwater styles of dictionary features toward the input.
	Moreover, a reference feature variant module (RFVM) is designed to adaptively morph the reference features, improving the similarity between the reference and input features. Experimental results on four UWI datasets show that our RFD-ECNet is the first work that achieves a significant BD-rate saving of 31$\%$ over the most advanced VVC. 
\end{abstract}


\vspace{-5pt}
\section{Introduction}

\begin{figure}[t]
	\centering
	\centerline{\includegraphics[width=1\linewidth]{fig1.pdf}}
	\vspace{-5pt}
	\caption{Key ideas of our work. Different UWIs have similarities in texture, structure and semantic, which can be included in an underwater feature dictionary to provide reference for UWI compression. E/D indicates encoder/decoder.}
	\label{figure1}
	\vspace{-10pt}
\end{figure}

Recently, underwater exploration attracts great attention from governments, scientists, and the public due to the growing underwater applications. For example, mine search, nuclear-reactors detection, and underwater power inspection \cite{Foresti2001} are parts of homeland security operations. In addition, marine biology \cite{Sundgren2003} and archaeology \cite{Kahanov2001} are important scientific research for ocean resource development. Moreover, underwater entertainment \cite{Chen2010} such as underwater live broadcasts is becoming greatly popular with the public.

In these underwater applications, images taken underwater play an essential role. To be transmitted from the deep sea to the board or ground, underwater images (UWIs) must be compressed at extremely low bitrates due to the very narrow underwater wireless acoustic bandwidth of about 20 kbps \cite{HallOCEANS2022}. However, existing image compression approaches severely degrade the pixel fidelity (\eg., MSE) of UWIs at extremely low bitrates ($<$0.1bpp), which greatly impairs the broad application of UWIs. Hence, it is highly urgent to design a more powerful extreme UWI compression algorithm.

Different from general images, UWIs present two unique underwater characteristics: 
\textbf{(1) UWIs present multifarious underwater styles of color shift and distance-dependent clarity.} 
This is because the underwater imaging process is quite different from that in the open air \cite{Chiang2012TIP}. When light travels through water, it suffers from absorption and scattering. Specifically, the blueish/greenish color shift is caused by the fact that the red light of the shortest wavelength is absorbed first, and then the green and blue light are followed \cite{Lin2021SPL}. In addition, light is scattered by the particles in water, which change the direction of light propagation \cite{Li2019TIP}, resulting in the spatially varying distance dependencies of clarity, $i.e.,$ the image clarity decreases as the distance to the objects increases.
\textbf{(2) Different UWIs contain some common underwater objects at diverse morphologies and sizes.} 
Due to the special underwater natural environment, there are some common underwater objects such as fish, corals, and rocks that widely exist in different UWIs in diverse morphologies and sizes. Hence, there are plenty of similar representations between UWIs. As shown in Fig. \ref{figure1} (b), although the UWIs are captured at different times and in various underwater scenes, they contain some same underwater objects such as water, corals, and rocks, which share plenty of similarities in textures, structures, and semantics.

Given the characteristics of UWIs, there are two main drawbacks for existing image compression methods \cite{JPEG2000, HEVC, VVC, Balle2018ICLR, Minnen2018NIPS, Zhu2022CVPR, 2021Li, Chen2021TIP, Cheng2020CVPR, ZouCVPR2022, Augstsson2019ICCV, Mentzer2020NIPS, Huang2019VCIP}: \textbf{(1) They only remove redundancy within an image and do not consider redundancy between UWIs.} Conventional image compression codecs such as JPEG2000 \cite{JPEG2000}, BPG \cite{HEVC} and the latest VVC-intra \cite{VVC} reduce the intra-redundancy through transform and intra-prediction. Recently, learnt image compression methods \cite{Balle2018ICLR, Minnen2018NIPS, Zhu2022CVPR, 2021Li, Chen2021TIP, Cheng2020CVPR, ZouCVPR2022} have shown greater potential than conventional codecs. Based on variational autoencoder (VAE), \cite{Balle2018ICLR} designs a hyperprior model to remove the statistic redundancy within an image. After that, \cite{Minnen2018NIPS} proposes an autoregressive model to further remove the local context redundancy. However, all of them are limited to removing redundancy in only an image. 
\textbf{(2) They utilize a unitary compression that cannot handle the multifarious underwater styles of color shift and distance-dependent clarity.} It is reasonable to use the unitary compression for the general terrestrial images because the colors (r, g, b) of terrestrial images are evenly distributed \cite{Ancuti2012} and the internal clarity is steady broadly. However, it limits their performance on UWIs because the UWIs are full of multifarious color shift and distance-dependent clarity. 
To design an efficient extreme UWI compression method, it needs to both remove redundancy between UWIs and align the multifarious underwater styles.

\begin{figure*}[t]
	\centering
	\includegraphics[width=1\linewidth]{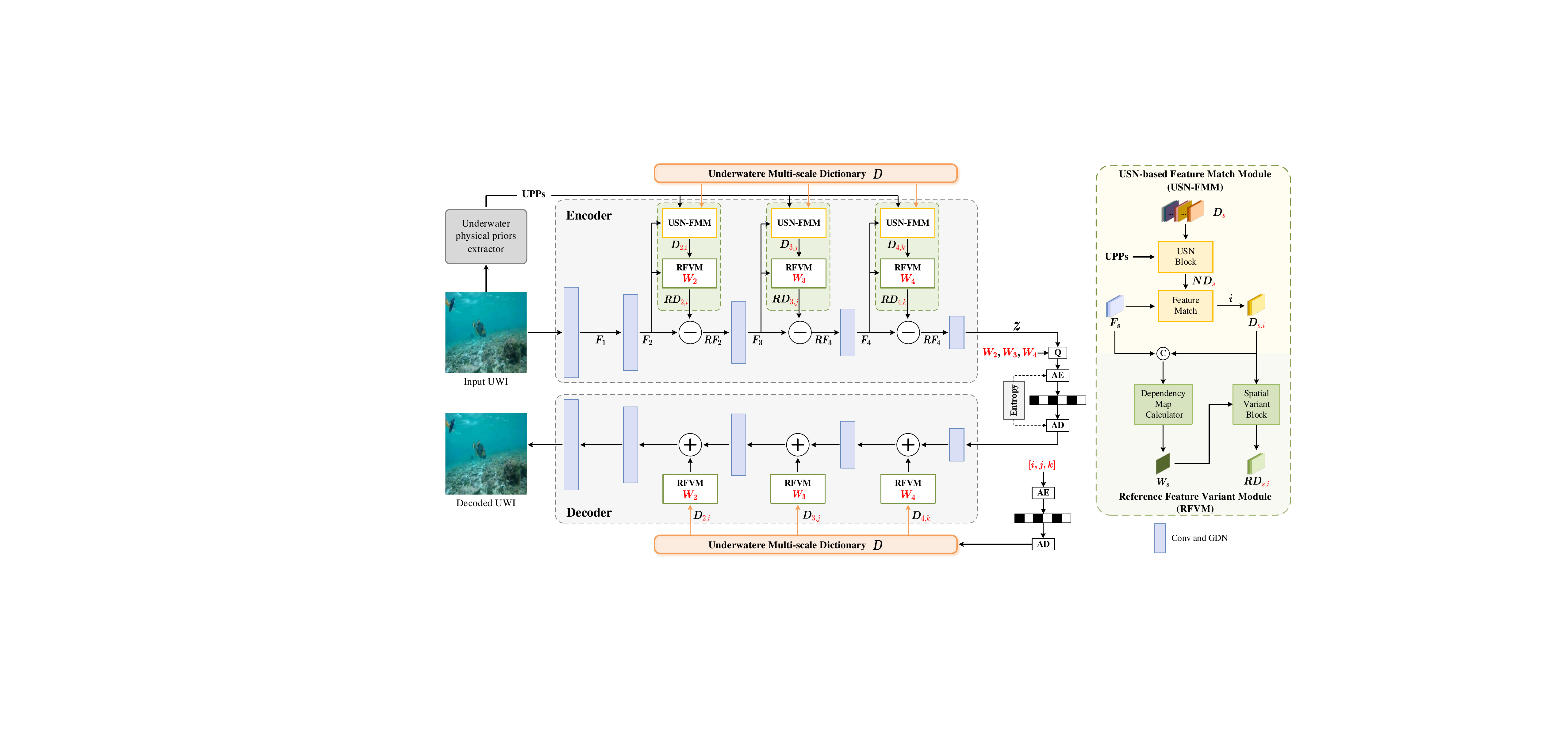}
	\vspace{-5pt}
	\caption{Overview of the proposed RFD-ECNet. Q indicates the quantizer. AE/AD represents the arithmetic en/decoding.}
	\label{figure2}
\end{figure*}

Accordingly, we creatively propose to compress UWIs by referencing a dictionary to remove redundancy between UWIs. First, we construct an exhaustive underwater multi-scale feature dictionary to provide from coarse to fine reference information for the UWI compression. Specifically, the features in the dictionary are extracted from plenty of representative UWIs, which are strictly selected according to various underwater scenes, including different underwater objects, water types, and water depths. As such, our dictionary could cover a wide range of underwater common objects, assisting the UWI compression to fully remove redundancy between UWIs. Subsequently, we design a novel extreme UWI compression network with reference to the underwater feature dictionary (RFD-ECNet), shown in Fig \ref{figure2}, where the features of input UWI are matched with the dictionary to select the reference features, and then the residuals between the input and reference features are compressed, significantly reducing the coding bits of UWIs.

Additionally, an underwater styles normalization block-based feature match (USN-FMM) is proposed to ensure that the match is not interfered by the multifarious underwater styles and focuses on selecting the references that have the most similar texture to the input. To be specific, the USNB utilizes the underwater physical priors (UPPs) extracted from the underwater imaging physical model \cite{Chiang2012TIP} to normalize the underwater style of reference features towards that of the input.
Moreover, to improve the similarity between the reference and input features, a reference feature variant module (RFVM) is designed to adaptively morph the reference features based on their dependency relevance to the input feature. Extensive experimental results show that our RFD-ECNet outperforms state-of-the-art (SOTA) compression methods. 
Overall, our main contributions are summarized as follows:
\begin{itemize}
	\vspace{-5pt}
	\item[$\bullet$] We make the first attempt to remove the massive redundancy between UWIs, and the proposed dictionary-based compression network RFD-ECNet is the first work that achieves significantly better BD-rate/PSNR than the latest VVC and other learnt methods.
	
	\vspace{-5pt}
	\item[$\bullet$] We construct an exhaustive underwater feature dictionary from representative UWIs manually selected based on various underwater scenes, water types, and water depths. Besides, detailed analyses are conducted to verify the comprehensiveness and compactness of the dictionary. 
	
	\vspace{-5pt}
	\item[$\bullet$] To eliminate the effect of underwater styles on feature match, we utilize the underwater physical priors to design USNB to improve feature match accuracy. Moreover, we design RFVM to adaptively morph the reference features based on the dependency map, improving their similarity.
\end{itemize}

\section{Related Works}
\textbf{Learnt Image Compression.} Recently, learnt image compression approaches have developed rapidly and achieved great breakthroughs, which can be divided into VAE-based and GAN-based. 
Generally, VAE-based methods \cite{Balle2018ICLR, Minnen2018NIPS, Zhu2022CVPR, 2021Li, Chen2021TIP, Cheng2020CVPR, ZouCVPR2022} utilize an encoder, which includes some parametric linear and nonlinear transforms, to compress image to compact latent features. After quantization and entropy coding, the latent features are compressed into bit-stream. Then a decoder, which usually has a symmetrical structure with the encoder, is used to reconstruct the image. 

To improve the rate-distortion (R-D) performance, some works focus on designing more efficient learnt model to capture redundancy in an image. \cite{Minnen2018NIPS} proposes auto-regressive model to combine the context model with the hyperprior model \cite{Balle2018ICLR} for entropy estimation. 
\cite{ZouCVPR2022} proposes window-based local attention block to capture local redundancy such as non-repetitive textures. 
\cite{Cheng2020CVPR} proposes Gaussian Mixture Model to estimate likelihoods of latent features more accurately.
Although VAE-based methods achieve admirable R-D performance, they often present unpleasing artifacts at extremely low bitrates. 

To this end, some works \cite{Augstsson2019ICCV, Mentzer2020NIPS, Huang2019VCIP} combine VAE with the generative adversarial network (GAN) to generate visually pleasing textures to counteract compression artifacts. Nevertheless, the pixel fidelity (\eg., MSE) of the image reconstructed by the GAN-based methods is quite low because the generated content is fake and deceptive. Hence, it is difficult for existing methods to achieve high pixel fidelity UWI reconstruction at extremely low bitrates by only removing redundancy within an image.

\textbf{Image Set Compression.}
This direction aims to compress a batch of images captured at the same place from different view angles and focal distances \cite{Sha2018ACCESS}, which can be divided into local-based and cloud-based. The local-based methods \cite{Zhang2018CSVT, Yeung2011ICME, MMD} construct the reference by first pre-processing all the images in a set. For example, \cite{Yeung2011ICME} creates a new low frequency template as reference image by averaging the low frequency components of all images. The cloud-based methods such as \cite{Zhao2015MMSP} select reference images from the massive images in the cloud.

Obviously, the local-based methods must obtain all images before compression, which limits their usage flexibility. The cloud-based methods rely on the wide transmission channel to access the cloud disk, which cannot be achieved by the very narrow underwater wireless acoustic channel. Moreover, since these methods can only be applied to the small image subset captured at the same place, they cannot deal with UWIs captured at different underwater scenes and depths in the vast underwater world.

\textbf{Reference-based Model.} To the best of our knowledge, there exists no image compression method based on image/feature reference. 
The closest area to our work is reference-based super-resolution (RefSR), which utilizes the high-frequency information in high-resolution (HR) reference image/patch to rich the details of low resolution (LR) image. Given an arbitrary HR image, \cite{Zhang2019CVPR} uses the VGG network to simultaneously extract the referenced HR and LR features, and swaps the most similar features of reference and the LR. To overcome the inherent information deficit of a single reference, \cite{Pesavento2021ICCV} uses multiple reference images to provide a more diverse pool of image features. Besides, \cite{Li2020ECCV} builds a face dictionary from HR face images to provide the high-frequency reference to construct more and richer features for the LR image.

The differences between our work and the RefSR works are as follows:
(1) The RefSR task aims to learn more/richer features from the HR reference, which is a process of increasing information. In contrast, our compression task aims to learn fewer/more compact features, a process of removing redundant information. To our best knowledge, we are the first image compression work referring to the feature dictionary. 
(2) In \cite{Li2020ECCV}, their dictionary only contains face images with eyes, nose and mouth, which are relatively more simple and similar than the vast UWI domain. Instead, UWIs contain multifarious underwater styles, where the common underwater objects are at diverse morphologies and sizes. To construct an exhaustive underwater dictionary, we manually select the representative UWIs based on various underwater scenes, water types, and water depths. 
(3) We design USNB to align the underwater styles of dictionary features and input, improving feature match accuracy. Additionally, we design RFVM to adaptively morph the reference features, further improving their similarity.

\section{The Proposed Method}
\subsection{Exhaustive Underwater Feature Dictionary}
\label{DFD}
\begin{figure}[t]
	\centering
	\includegraphics[width=1\linewidth]{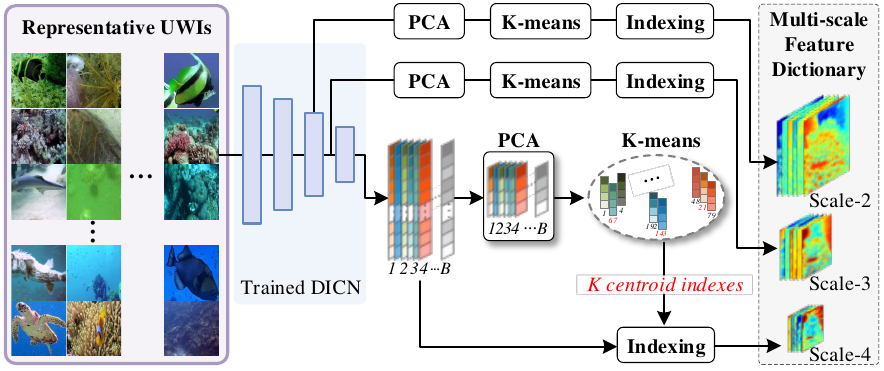}
	\vspace{-15pt}
	\caption{Construction of underwater multi-scale feature dictionary. $B$ is the number of the representative UWIs.}
	\label{figure3}
	\vspace{-5pt}
\end{figure}

To achieve extremely low bitrate UWI compression, we propose RFD-ECNet, a novel image compression network referring to underwater feature dictionary, to remove redundancy between UWIs. First of all, an exhaustive dictionary is constructed, which is offline and shared at the encoder and decoder. The pipeline of the construction of underwater multi-scale feature dictionary is shown in Fig. \ref{figure3}.

To build the feature dictionary, a series of representative UWIs are selected from the largest public underwater image dataset UGWI \footnote{UGWI is a public underwater image dataset and is available online at \url{https://github.com/Underwater-Lab-SHU}}. UGWI contains 2150 various UWIs, which are collected during oceanic explorations, human-robot collaborative experiments, or from websites such as Google and YouTube. To select representative UWIs that cover the richest and most comprehensive common contents of UWIs, we manually divided the two thousand images into 300 groups according to two rules.
\textbf{1) Different Objects.} Based on \cite{Li2019TIP}, the common objects in UWIs can be divided into marine faunas of fish, turtles, sharks, snails and shellfish, \emph{etc.}; marine plants of corals, seaweed, fringing reefs, \emph{etc.}; non-biology of rocks, wreckage and sculpture, \emph{etc.}
\textbf{2) Different water types and depths.} The water type is distinguished from clear to turbid, and the water depth is distinguished from shoal water, deep sea and seabed. UWIs captured in clear and shoal water present lighter color shift and better clarity, and vice versa. 

\begin{figure}[t]
	\centering
	\includegraphics[width=0.5\linewidth]{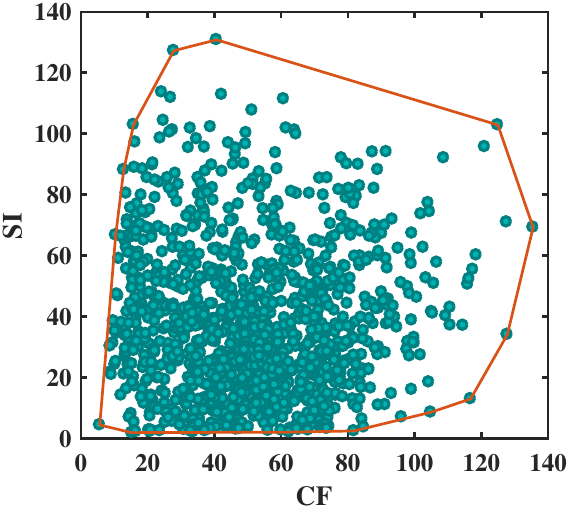}
	\vspace{-5pt}
	\caption{SI and CF of images used to build the dictionary.}
	\label{figure4}
	\vspace{-5pt}
\end{figure}

After grouping the UGWI manually, 1-2 images from each group are selected to build our dictionary, which is then cropped into non-overlapping 128$\times$128 image patches $P$. For each image patch, we use the pre-trained popular deep image compression network DICN \cite{Balle2018ICLR} to extract its compression features at multi-scales, shown in Fig. \ref{figure3}. The generation process of dictionary can be described as,
\begin{equation}
	D_s = \mathcal{F}(P;\theta_{DICN}^s)
	\label{eq_1}
\end{equation}
where $s \in \{2,3,4\}$ is the scale and $\theta_{DICN}^s$ is the parameters of the pre-trained DICN at scale $s$.

To verify the diversity of these representative UWIs, the Spatial Information (SI) and Colorfulness (CF) indices are computed respectively for the content variation along the spatial and color dimensions. Higher SI / CF indicates more complex image content / colorful. The CF versus SI distribution is plotted in Fig. \ref{figure4}. As shown, the distribution range is quite broad, which indicates that the content of these patches is diverse. Additionally, some points are distributed closely, which also verifies that there are many similar common content between UWIs.

To compact the features in dictionary, we adopt K-means to create $K$ clusters. Considering the high computational complexity of directly performing K-means on massive features of multifarious UWIs, we first perform principal components analysis (PCA) on the features of per UWI to reduce their dimension. 
Finally, $K$ typical features at each scale are preserved, constructing our multi-scale feature dictionary. The effect of cluster number $K$ on the compression performance is shown in the supplementary materials.

\subsection{Overall of the proposed RFD-ECNet}
The framework of our RFD-ECNet is shown in Fig. \ref{figure2}, which performs feature match at multiple scales ($s \in \{2, 3, 4\}$) in a progressive manner to remove the underwater common redundancy from coarse to fine. 
Let $F_s$ be the input features at scale $s$, and $D_{s,i}$ denotes the $i$-$th$ dictionary features at scale $s$. In the encoder, the input UWI is down-sampled by two convolutional sampling layers to obtain its features $F_2$, which are matched with the dictionary $D$ to select its reference features $D_{2,i}$.  
To guarantee the input features $F_2$ and $D$ in the same feature space, the first-two sampling layers of RFD-ECNet are filled with that of the pre-trained DICN, which are fixed during the training of our RFD-ECNet. 

For more accurate feature match, USNB is proposed to normalize the dictionary features based on the underwater style of the input, eliminating the effect of the underwater styles diversity on feature match. Specifically, USNB utilizes the UPPs extracted from an UPPs extractor \cite{Lin2021SPL} to simulate the underwater imaging process \cite{Chiang2012TIP}, which dictate the underwater styles of UWIs. Moreover, the selected $D_{2,i}$ is further refined by the proposed RFVM to obtain $RD_{2,i}$ based on the similarity map $W_2$ between $F_{2}$ and $D_{2,i}$, greatly improving their similarity. 

After removing redundancy on three scales, the latent features $z$ and the dependency maps ($W_2, W_3, W_4$) are quantized by the uniform noise approximated method \cite{Balle2018ICLR}, and then entropy coded by the Gaussian mixture entropy model \cite{Cheng2020CVPR} to the bit-stream. 

\begin{figure}[t]
	\centering
	\includegraphics[width=1\linewidth]{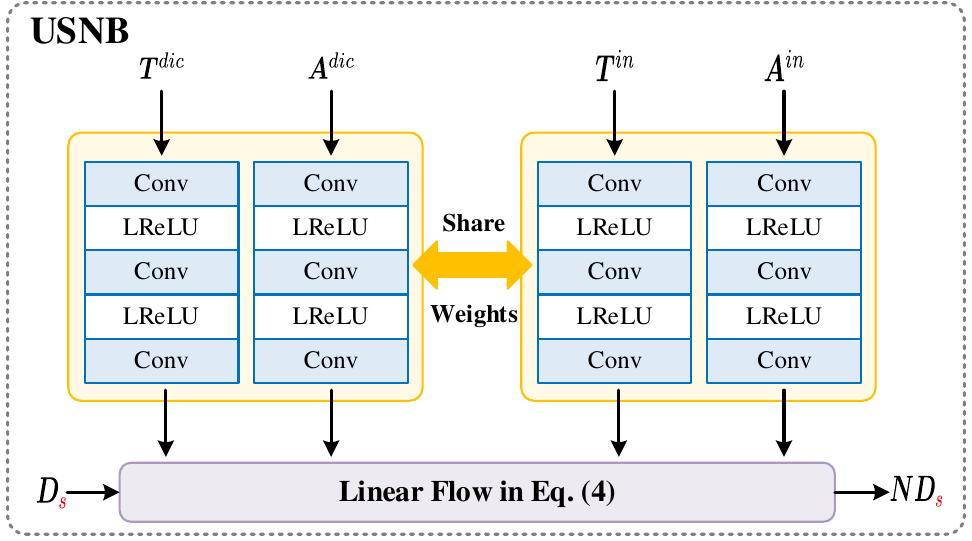}
	\vspace{-15pt}
	\caption{The structure of the proposed USNB.}
	\label{figure5}
	\vspace{-5pt}
\end{figure}

\subsection{USNB-based Feature Match (USN-FMM)}
\label{USNB}
Feature match aims to select the reference features that have the most similar textures to the input features. However, the UWIs sharing similarity may have multifarious underwater styles of color shift and distance-dependent clarity. Hence, it is necessary to normalize the underwater styles of dictionary features toward that of the input before the feature match. 

\textbf{Underwater Style Normalize.}
By studying the process of underwater imaging, it can be observed that the underwater styles are dictated by underwater physical priors such as ambient light and imaging transmission map, which can be described by the underwater physical scattering model:
\begin{equation}
	\begin{aligned}
		J &= I \times T + A \times (1-T) \\
		T &= e^{-\alpha d}
	\end{aligned}
	\label{eq2}
\end{equation}
where $I$ denotes the clear scene without underwater style and $J$ is the captured UWI with multifarious underwater styles. Under the effect of underwater physical priors, including $T$ and $A$, $I$ is evolved to $J$. $T$ and $A$ respectively represent the transmission map and the global ambient light, and their visual examples are presented in the supplementary materials. $T$ is calculated by $\alpha$ and $d$, which respectively refer to light attenuation coefficient and the distance between camera and object, which leads to the distance-dependent clarity. Derived from Eq. (\ref{eq2}), the clear scene can be modeled as: 
\begin{equation}
	I = \frac{J-A\times (1-T)}{T}
	\label{eq3}
\end{equation}
Based on Eq. (\ref{eq2}) and Eq. (\ref{eq3}), the dictionary features can be normalized towards the underwater style of input by 
\begin{equation}
	\begin{split}
		J^{dic\_input} &= I^{dic}\times T^{input} + A^{input} \times (1-T^{input}) \\
		&= \frac{J^{dic}-A^{dic}\times (1-T^{dic})}{T^{dic}} \times T^{input} \\
		&+A^{input} \times (1-T^{input})
	\end{split}
	\label{eq4}
\end{equation}

\begin{figure}[t]
	\centering
	\includegraphics[width=1\linewidth]{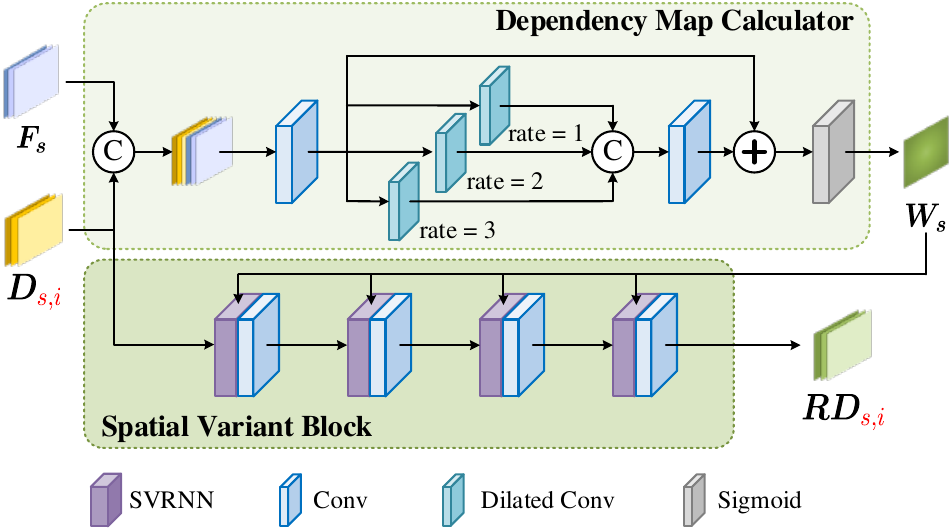}
	\vspace{-15pt}
	\caption{The structure of the proposed RFVM.}
	\vspace{-5pt}
	\label{figure6}
\end{figure}
\begin{figure*}[t]
	\centering
	\begin{subfigure}{0.246\linewidth}
		\includegraphics[width=1\linewidth]{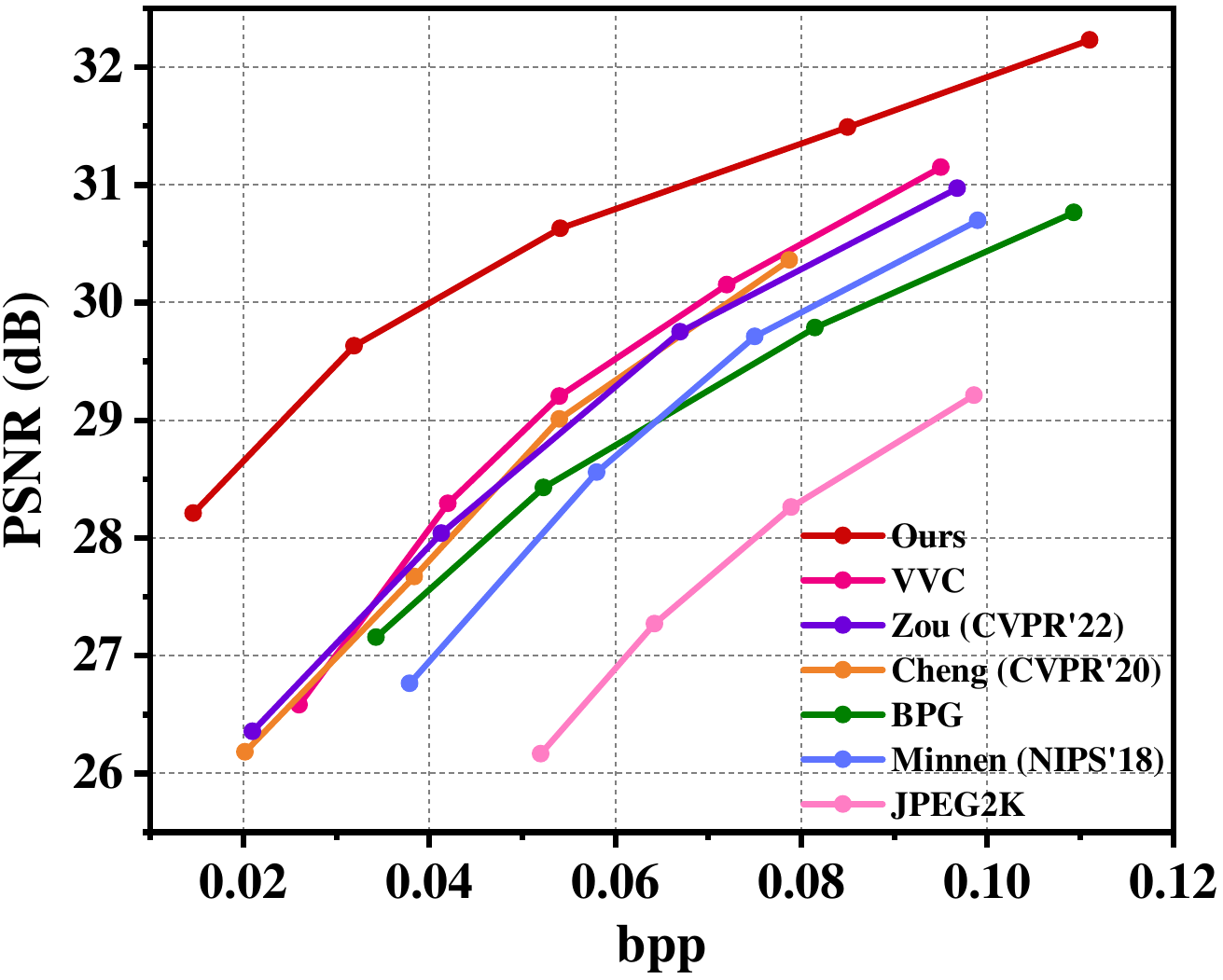}
		\vspace{-15pt}
		\caption{UGWI, Rate-PSNR}
	\end{subfigure}
	\hfill
	\begin{subfigure}{0.246\linewidth}
		\includegraphics[width=1\linewidth]{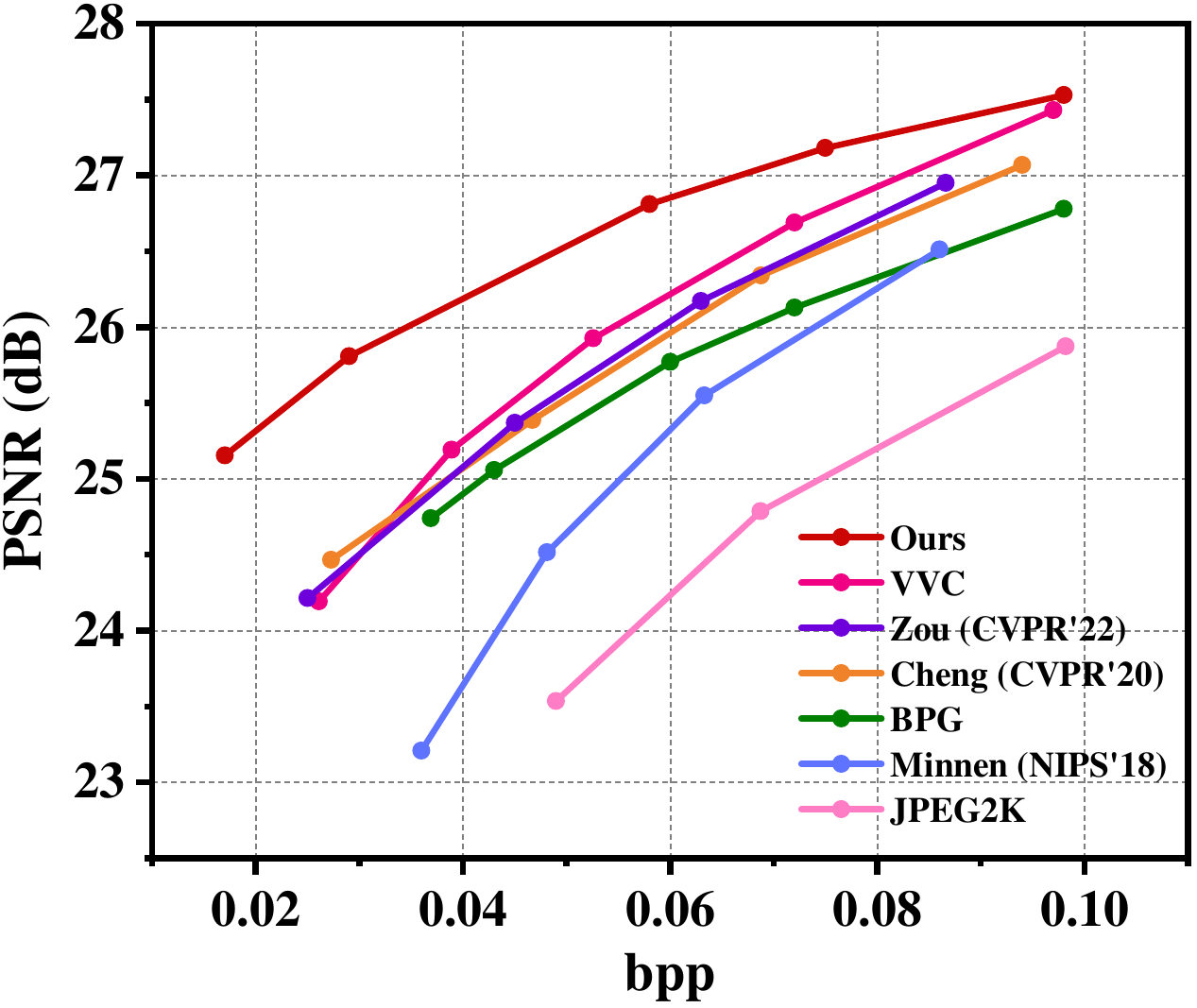}
		\vspace{-15pt}
		\caption{EUVP, Rate-PSNR}
	\end{subfigure}
	\hfill
	\begin{subfigure}{0.246\linewidth}
		\includegraphics[width=1\linewidth]{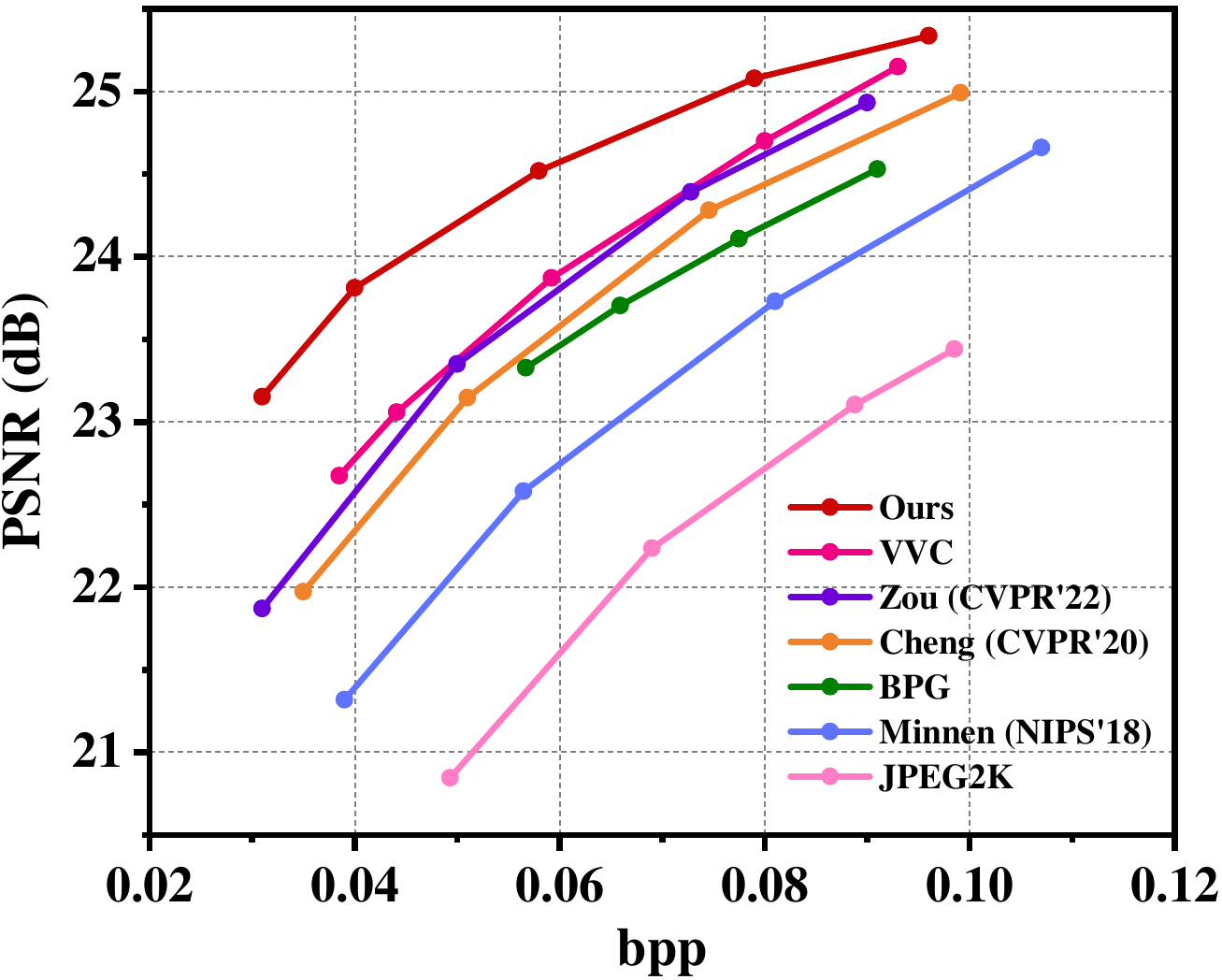}
		\vspace{-15pt}
		\caption{UFO, Rate-PSNR}
	\end{subfigure}
	\hfill
	\begin{subfigure}{0.246\linewidth}
		\includegraphics[width=1\linewidth]{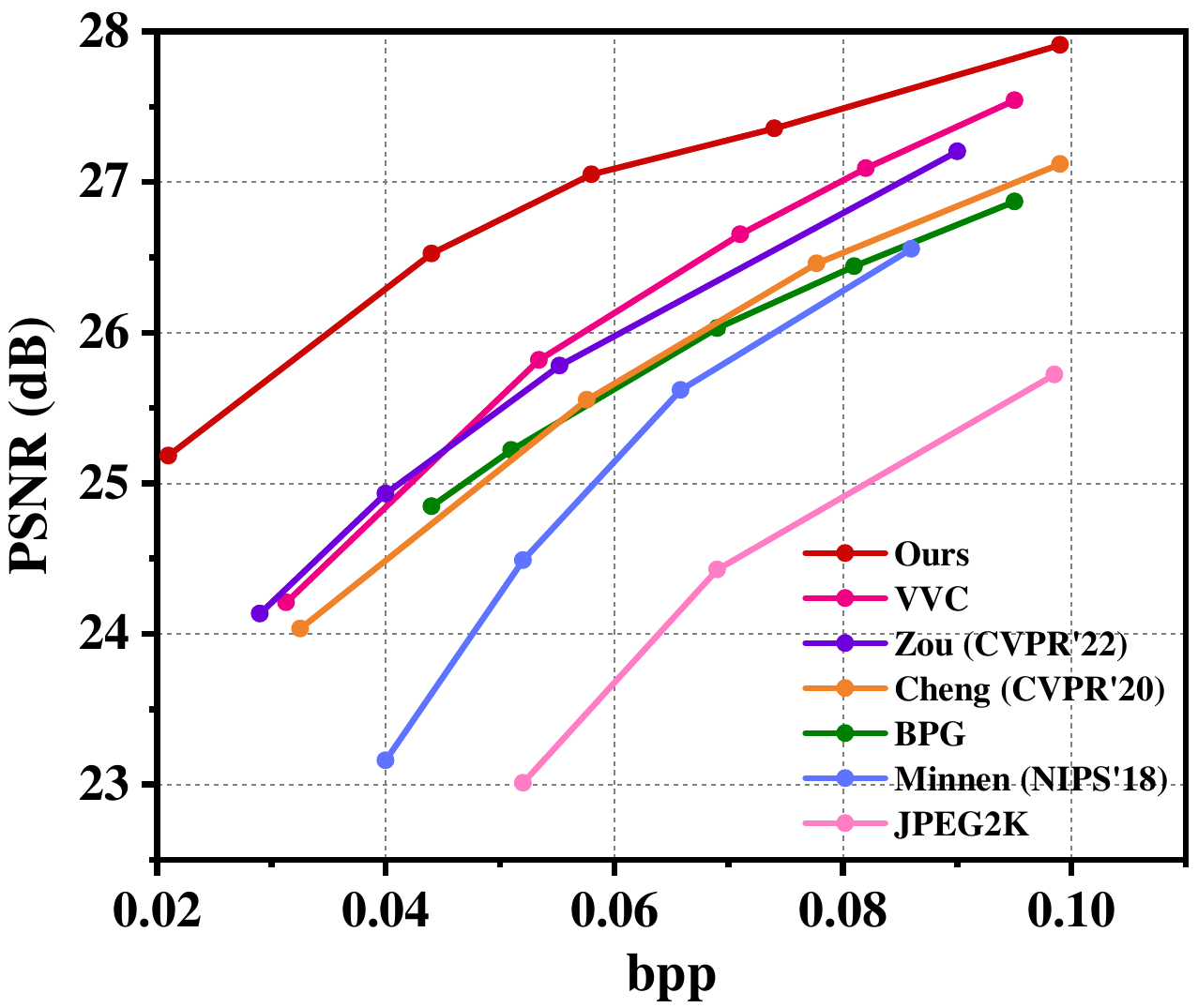}
		\vspace{-15pt}
		\caption{UIEB, Rate-PSNR}
	\end{subfigure}
	\vspace{-5pt}
	\caption{R-D curves of different image compression methods on UGWI, EUVP \cite{EUVP}, UFO \cite{UFO}, and UIEB \cite{Li2019TIP} datasets.}
	\label{figure7}
	\vspace{-5pt}
\end{figure*}

Based on the above analysis, we propose the USNB to parametrically implement Eq. (\ref{eq4}), as shown in Fig. \ref{figure5}, which is used before the feature match to normalize the underwater styles of dictionary features toward the input. In Fig. \ref{figure5}, $D_s$ and $ND_s$ respectively correspond to $J_{dic}$ and $J_{dic\_input}$ in Eq. (\ref{eq4}). Underwater physical priors of dictionary features ($T^{dic}$ and $A^{dic}$) and that of the input ($T^{in}$ and $A^{in}$) are fed to several convention layers to normalize $D_s$ to $ND_s$ by following the data flow in Eq. (\ref{eq4}).
Notably, the network parameters for processing $T^{dic}$ and $A^{dic}$ are shared with that for $T^{dic}$ and $A^{dic}$ to guarantee that all underwater physical priors are in the same feature space. As such, the proposed USNB is able to normalize the underwater styles of the dictionary features toward the input for more accurate feature match. The effectiveness of USNB is verified in the ablation studies.

\textbf{Feature Match.} To select the reference features $D_{s,i}$ from the dictionary features $D_s$ for input features $F_s$, we adopt the widely used inner product to measure the similarity score between $F_s$ and $D_s$. The similarity scores between $F_s$ and $D_s$ is defined as:
\vspace{-8pt}
\begin{equation}
	\vspace{-3pt}
	S_{s} = \left \langle F_s, \frac{D_{s}}{\left \|D_{s}\right \|} \right \rangle
	\label{eq5}
\end{equation}
Following the procedure in \cite{2019CVPR}, Eq. (\ref{eq5}) is implemented by a convolution operation, where the kernel weights are set as $D_s$, to accelerate the similarity calculations. The $i$-th dictionary feature with the highest similarity score is selected as the reference feature $D_{s,i}$.

\subsection{Reference Feature Variant Module (RFVM)}
\label{RFVM}
Although the reference features $D_{s,i}$ and the input features $F_s$ share some similarities, the relative position and shape of the similar areas in $D_{s,i}$ and $F_s$ may be dramatically different. To achieve the highest possible similarity, we develop the RFVM to adaptively morph the $D_{s,i}$ based on $F_s$, shown in Fig. \ref{figure6}. RFVM utilizes the spatial variant recursive convolution (SVRConv) \cite{Zhang2018CVPR} to implement the feature variant on a large receptive field, which is able to facilitate the dramatic spatial variant by the long-distance information propagation of recurrent convolution and the learned weights for different locations. The weights of SVRConv are learned by exploring the dependency between $D_{s,i}$ and $F_s$. To obtain the global dependency map $W$, dilated convolutions with different dilation rates are used to exploit the spatial information at different receptive fields. As such, our RFVM is able to morph the reference features in a large receptive field to improve the similarity with the input features.

\section{Experiments}
\label{sec:Experiments}
Extensive experiments are conducted to evaluate the extreme compression performance of our RFD-ECNet. First, we test RFD-ECNet on four different UWI datasets objectively and subjectively to verify our excellent generalization. After that, a series of ablation studies are presented to respectively verify the effectiveness of our multi-scale manner, USNB and RFVM.

\begin{table*}[t]
	\caption{Comparison of BD-rate and BD-PSNR on UGWI, EUVP \cite{EUVP}, UIEB \cite{Li2019TIP}, and UFO \cite{UFO} testsets. The BD-rate and BD-PSNR are calculated from PSNR-BPP curve over data points with bpp$\textless$0.1, when BPG \cite{HEVC} is set as the anchor. The performance of SOTA compression codec VVC is marked in \textcolor{red}{red}. The best performance is \textbf{bolded}.} 
	\vspace{-15pt}
	\begin{center}
		\resizebox{1\linewidth}{!}{
			\renewcommand{\arraystretch}{1.2}
			\begin{tabular}{c|cccc|c|cccc|c|cc}
				\hline \hline
				\multirow{2}*{Methods} & \multicolumn{5}{c}{BD-rate $\downarrow$ (\%)} \vline &\multicolumn{5}{c}{BD-PSNR $\uparrow$ (dB)} \vline &\multicolumn{2}{c}{\makecell [c]{Computation complexity}}\\ \cmidrule(r){2-13}
				&UGWI & EUVP &UIEB &UFO &$Avg.$ &UGWI & EUVP &UIEB &UFO &$Avg.$ &Param. (M) & FLOPs (G)\\
				\midrule
				\textcolor{red}{VVC-intra} \cite{VVC} &\textcolor{red}{-19.0}  &\textcolor{red}{-16.2}   &\textcolor{red}{-16.0} &\textcolor{red}{-16.1} &\textcolor{red}{-16.8} &\textcolor{red}{0.69} &\textcolor{red}{0.43} &\textcolor{red}{0.52} &\textcolor{red}{0.47} &\textcolor{red}{0.53} &/ &/\\
				JPEG2000 \cite{JPEG2000}&61.7 &68.4 &67.2 &66.9 &66.1 &-1.62 &-1.71 &-1.84 &-1.56 &-1.68 &/ &/\\
				BPG \cite{HEVC} &0 &0 &0 &0 &0 &0 &0 &0 &0 &0 &/ &/\\
				Minnen (NIPS'18) \cite{Minnen2018NIPS} &-3.8 &14.8 &9.8 &19.0  &9.9 &0.16 &-0.36 &-0.33 &-0.59 &-0.28 &14.1 &27.2\\
				Cheng (CVPR'20) \cite{Cheng2020CVPR}&-12.8 &-8.4 &-1.3 &-8.3 &-7.7 &0.43 &0.20 &0.04 &0.24 &0.23 &20.5 &61.3\\
				Zou (CVPR'22) \cite{ZouCVPR2022}&-13.8 &-10.3 &-12.4 &-14.7 &-12.8 &0.48 &0.25 &0.36 &0.43 &0.38 &75.0 & 33.3\\ \hline
				Ours-slim &\textbf{-50.2} &\textbf{-44.6} &\textbf{-42.6} &\textbf{-34.7} &\textbf{-43.0} &\textbf{1.73} &\textbf{1.01} &\textbf{1.11} &\textbf{0.65} &\textbf{1.12} &15.4 &40.7 \\
				Ours &\textbf{-53.7} &\textbf{-50.3} &\textbf{-52.9} &\textbf{-37.3} &\textbf{-48.5} &\textbf{1.97} &\textbf{1.18} &\textbf{1.56} &\textbf{0.99} &\textbf{1.43} &35.8 &86.7\\
				\hline \hline
			\end{tabular}
			\label{table1}}
	\end{center}
\end{table*}

\begin{figure*}[t]
	\vspace{-10pt}
	\centering
	\centerline{\includegraphics[width=1\linewidth]{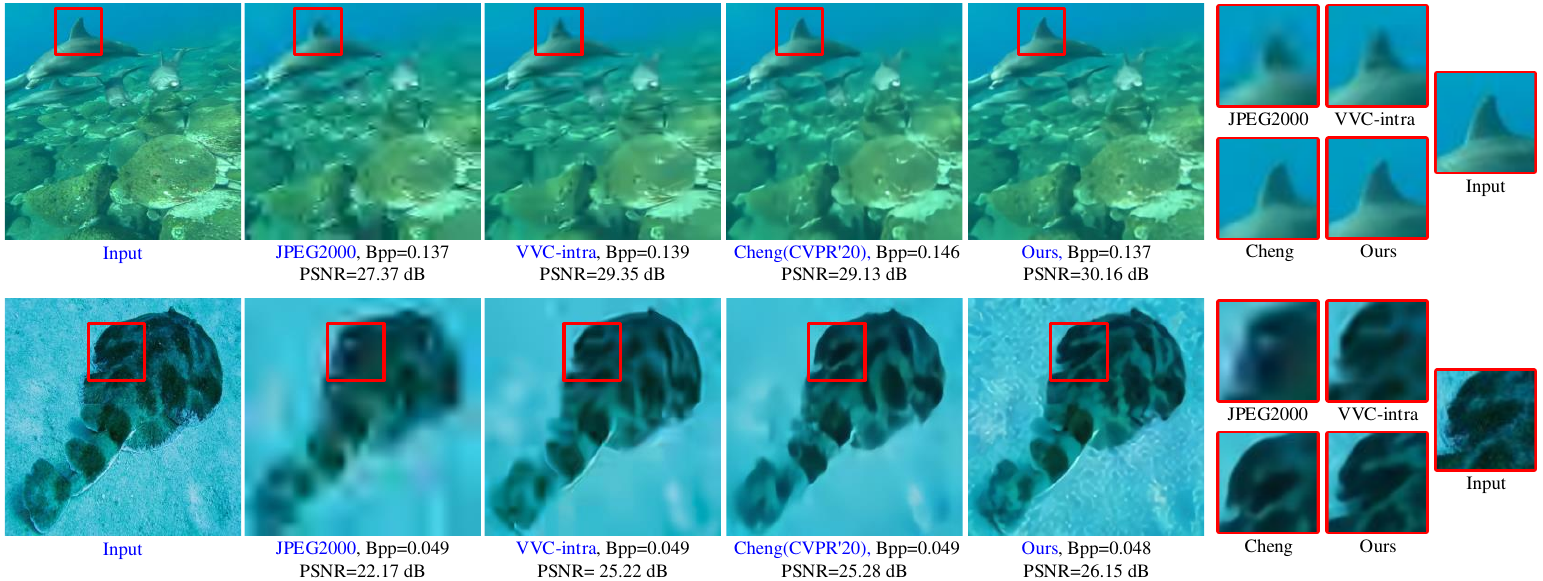}}
	\vspace{-5pt}
	\caption{Visual examples of the reconstructed images. The examples in two rows are respectively from UGWI and EUVP \cite{EUVP} testset. The metrics are bpp ($\downarrow$) and PSNR ($\uparrow$). At extremely low bpp, our reconstructed images have more details and higher pixel fidelity (measured by PSNR) than others.}
	\label{figure8}
\end{figure*}

\subsection{Experimental Settings}
\label{sec:settings}
{\bf Datasets.} We perform experiments on four public UWI datasets, including the UGWI dataset, EUVP dataset \cite{EUVP}, UFO dataset \cite{UFO}, and UIEB dataset \cite{Li2019TIP}. These UWI datasets are collected during different oceanic explorations and hence include various underwater scenes and rich ocean species. As stated in Section \ref{DFD}, some representative UWIs selected from UGWI dataset are utilized to construct our underwater multi-scale feature dictionary. The rest of UGWI are divided into 1550 UWIs for training and 300 UWIs for testing. Additionally, we randomly select 100 images respectively from EUVP \cite{EUVP}, UFO \cite{UFO}, and UIEB \cite{Li2019TIP} to test the pre-trained RFD-ECNet to verify the generalization of our RFD-ECNet.

{\bf Training.} As the most VAE-based methods, our RFD-ECNet is optimized by the rate-distortion trade-off loss function, which can be described as:
\begin{equation}
	\vspace{-5pt}
	\mathcal{L} = \lambda \cdot D + R,
\end{equation}
where $\lambda$ is a trade-off parameter to balance bitrate $R$ and distortion $D$. In our work, MSE is used as the $D$, and $\lambda$ belongs to the set \{512, 256, 128, 64, 32\} for 5 different compression ratios. The channel width of RFD-ECNet is set as 192, and the bottleneck layer is set as 64. Detailed network architecture is introduced in the the supplementary materials. All model are trained using the Adam optimizer \cite{adam} with standard parameters and learning rate of $1\times 10^{-4}$.

{\bf Evaluation.} The compression performance is measured by bitrate $R$ and distortion $D$, which are respectively calculated by bits-per-pixel (bpp) and Peak Signal to Noise Ratio (PSNR). Lower bpp and higher PSNR indicate better compression performance. Notably, our bpp is calculated by, 
\begin{equation}
	bpp = \frac{bits_{z} + bits_{W} + bits_{index}}{N_{pixels}}
	\label{eq7}
\end{equation}
where $bits_{z}$ and $bits_{W}$ respectively represent the bits of latent features $z$ and dependency map $W$. $bits_{index}$ denotes the bits for coding the index numbers of reference features, which are directly coded by Huffman coding. In addition to R-D curves, the widely used Bjøntegaard delta rate (BD-rate) and BD-PSNR are also evaluated. Lower BD-rate and higher BD-PSNR indicate more efficient compression.

For the comparative methods, we compare our RFD-ECNet with influential and SOTA image compression networks, including Minnen (NIPS'18) \cite{Minnen2018NIPS}, Cheng (CVPR’20) \cite{Cheng2020CVPR}, and Zou (CVPR’22) \cite{ZouCVPR2022}. For fair comparisons, all networks are trained on the UGWI dataset. Additionally, some conventional codecs are compared, including popular JPEG2000 \cite{JPEG2000}, BPG (HEVC-intra) \cite{HEVC} and the latest VVC-intra (VTM 11.0) \cite{VVC}. For BPG and VVC, the highest PSNR setting (4:4:4) is tested.

\subsection{Comparison against SOTA methods}
{\bf R-D Performance.} Fig. \ref{figure7} presents the R-D curves on four UWI testsets of UGWI, EUVP \cite{EUVP}, UIEB \cite{Li2019TIP}, and UFO \cite{UFO}. As shown, the R-D curves of our RFD-ECNet are above the curves of all comparative methods on four testsets, which demonstrates that our method achieves better compression performance than all comparative methods and our excellent generalization and robustness. Notably, as bpp decreases, our RFD-ECNet achieves higher PSNR improvements than comparative methods. This indicates that the reference features provided by our underwater feature dictionary are very efficient for the reconstruction, especially at extremely low bitrates. 

\vspace{-3pt}
The BD-rate of different image compression methods with BPG \cite{HEVC} as the  anchor is presented in Table \ref{table1}. As shown, our method achieves the best BD-rate and BD-PSNR results on four testsets, which indicates that we use the smallest amount of bits to achieve the highest PSNR among all methods. Specially, our average BD-rate of -48.5 $\%$ and BD-PSNR of 1.43 dB are much better than the most advanced codec VVC-intra' -16.8 $\%$ and 0.53 dB. More intuitively, when VVC is set as the anchor, our RFD-ECNet achieves breakthrough BD-rate saving of 31.7 $\%$ and BD-PSNR of 0.90 dB, demonstrating the advanced performance of our RFD-ECNet on UWI extreme compression.

{\bf Visual Quality.}
Fig. \ref{figure8} shows visual examples of reconstructed images by our RFD-ECNet, Cheng (CVPR'20), the most advanced VVC-intra, and the widely used JPEG2000. As shown, our reconstructed images retain more details, while the reconstructed images of VVC-intra and JPEG2000 present obvious blurring and blocking effects at extremely low bpp. As shown in the second row, when bpp=0.049, VVC-intra, Cheng (CVPR'20) and JPEG2000 completely loss the common object of sand, while our RFD-ECNet can roughly reconstruct the sand by referencing our dictionary. Overall, our pixel fidelity is higher than others at the similar extremely low bpp, which can be verified by our higher PSNR. More visual results tested on underwater videos are provided at \url{https://github.com/lilala0/RFD-ECNet}.

{\bf Comparison with video codecs.}  To verify that the redundancy between UWIs cannot be efficiently removed by video codecs, our RFD-ECNet is compared with the latest video codecs of VVC-inter and HEVC-inter. Their performance on UGWI testset measured by BD-rate and BD-PSNR with the Minnen \cite{Minnen2018NIPS} as the anchor is shown in Table \ref{table2}.
As shown, we achieve better results than both VVC-inter and HEVC-inter, verifying that our RFD-ECNet referencing to dictionary is more efficient than video codecs to remove the redundancy between UWIs. 
Moreover, both video codecs at inter-mode only achieve tiny improvement of BD-rate/PSNR than the intra-mode. This verifies that there is redundancy between UWIs indeed.
\begin{table}[t]
	\caption{BD-rate ($\%$) and BD-PSNR (dB) of VVC/HEVC at inter/intra modes, where Minnen \cite{Minnen2018NIPS} is set as the anchor.}
	\vspace{-15pt}
	\begin{center}
		\resizebox{1\linewidth}{!}{
			\renewcommand{\arraystretch}{1.2}
			\centering
			\begin{tabular}{l|cc|cc|c}
				\hline \hline
				Methods &VVC-inter &VVC-intra &HEVC-inter &HEVC-intra &Ours \\
				\midrule
				BD-rate $\downarrow$ &-22.1 &-21.0 &0.2 &3.2 &\textbf{-59.5} \\
				BD-PSNR $\uparrow$ &0.85 &0.82 &-0.10 &-0.12 &\textbf{2.25}  \\
				\hline \hline
		\end{tabular}	}
	\end{center}
	\vspace{-20pt}
	\label{table2}
\end{table}

{\bf Comparison of complexity.} The trainable parameters and FLOPs of all approaches are compared in Table \ref{table1}. 
As shown, our parameters (35.8 M) is much less than that (99.8 M) of compression network \cite{ZouCVPR2022}. Additionally, we also provide a slim version of Our RFD-ECNet by narrowing our network channels of 192 to 128 (denoted as Ours-slim), which only have 15.4M parameters and 40.7G FLOPs. Notably, ours-slim still achieves better performance than all SOTA methods with lower complexity, illustrating our excellent UWI extreme compression performance.


\section{Ablation Study}
\label{Ablations}
{\bf Effectiveness of USNB and RFVM.}
To verify the effectiveness of the proposed USNB and RFVM, a series of ablation experiments are conducted, where the USNB and RFVM are removed from our RFD-ECNet in sequence. The performance of each ablation is measured by the BD-rate and BD-PSNR with the BPG as the anchor, shown in Table \ref{table3}. \emph{w/o USNB} indicates the network that directly performs feature match by inner product without using our USNB to normalize the dictionary features. \emph{w/o RFVM} is the network where the matched reference features are not fed into our RFVM and therefore cannot be varied adaptively based on the input features. \emph{w/o USNB\&RFVM} is the network without both USNB and RFVM. As can be seen from Table \ref{table3}, the removal of either USNB or RFVM degrades the performance of our RFD-ECNet. Moreover, the performance degrades most when USNB and RFVM are removed simultaneously, verifying the effectiveness of USNB and RFVM.

{\bf Effectiveness of multi-scale reference.} In our RFD-ECNet, the input features are matched with the feature dictionary at three scales ($s=2, 3, 4$), considering that the similarity between UWIs is reflected on the textures, structures, and semantics. Here we conduct ablations to verify the effectiveness of the multi-scale manner, and provide the results in Table \ref{table4}. Since the features on scale-1 contain more noise less semantic information, we perform feature match starting from scale-2.
As shown, our RFD-ECNet referring dictionary at scales-\{$2, 3, 4$\} achieves the best performance, followed by scales-\{$2, 3$\} and scale-\{$2$\}, which verifies that it is efficient to remove the underwater common redundancy at multiple scales from coarse to fine.

\begin{table} [t]
	\caption{BD-rate ($\%$) and BD-PSNR (dB) of ablation studies about our USNB and RFVM, where BPG \cite{HEVC} is set as the anchor.}
	\vspace{-15pt}
	\begin{center}
		\resizebox{1\linewidth}{!}{
			\renewcommand{\arraystretch}{1.2}
			\centering
			\begin{tabular}{l|cccc}
				\hline \hline
				Setting &Ours & w/o USNB & w/o RFVM & w/o USNB\&RFVM \\
				\midrule
				BD-rate $\downarrow$ &-53.7 &-33.6 &-43.1 &-24.6  \\
				BD-PSNR $\uparrow$ &1.97 &1.14 &1.49 &0.83 \\
				\hline \hline
		\end{tabular} }
	\end{center}
	\label{table3}
\end{table}
\begin{table} [t]
	\vspace{-15pt}
	\caption{BD-rate ($\%$) and BD-PSNR (dB) of ablation studies about the multi-scale manner, where BPG \cite{HEVC} is set as the anchor.}
	\vspace{-15pt}
	\begin{center}
		\resizebox{1\linewidth}{!}{
			\renewcommand{\arraystretch}{1.2}
			\centering
			\begin{tabular}{l|cccc}
				\hline \hline
				Scale setting &~~~Scale-\{2\}~~~ &~~Scales-\{2, 3\}~~ &~~~Scales-\{2, 3, 4\}~~~ \\ \hline
				BD-rate $\downarrow$ &-46.1 &-49.8 &-53.7 \\
				BD-PSNR $\uparrow$ &1.67 &1.83 &1.97\\
				\hline \hline
		\end{tabular} }
	\end{center}
	\label{table4}
	\vspace{-15pt}
\end{table}


\section{Discussion}
This work creatively removes redundancy between UWIs and proposes a novel reference-based image compression framework. Actually, the redundancy between images exists not only in UWIs but also in other image domains where independent images contain some common objects specific to the image domain. For example, all Martian images \cite{Ding2022VCIP} contain rocks and sand; all face images \cite{Cao2018FG} contain eyes, nose and mouth; most remote sensing images \cite{Xia2010ISPRS} contain buildings, road and so on. However, existing image compression works neglect this nature and only remove redundancy in an image, which may limit the development of image compression.

To validate our hypothesis, we additionally conduct preliminary experiments on Martian images \cite{Ding2022VCIP} by adjusting our network. Experimental results presented in the supplementary materials show that we also achieve better performance than VVC and Ding (VCIP'22) \cite{Ding2022VCIP}, illustrating the efficiency and feasibility of removing redundancy between images in other image domains.



\section{Conclusion}
In this paper, we explore the characteristics of UWIs and find that UWIs present multifarious underwater styles and different UWIs share some common ocean objects at diverse morphologies and sizes, resulting in plenty of redundancy between UWIs.
Hence, we construct an exhaustive and compact underwater multi-scale feature dictionary from the manually selected representative UWIs, which provides coarse-to-fine reference for our RFD-ECNet to fully remove redundancy between UWIs. Meanwhile, to eliminate the effect of underwater styles on feature match, we utilize UPPs to design the USNB to align the multifarious underwater styles of dictionary towards that of input features. Moreover, to improve the similarity between the input and reference features, we propose RFVM to perform adaptive reference feature variant. Finally, extensive comparisons are conducted to verify our significant performance improvement than SOTA methods including the latest VVC, and comprehensive ablation studies verify the effectiveness of each module in our RFD-ECNet.
\\ \\{\large{\textbf{Acknowledgments}}}\\

This work is supported in part by the National Natural Science Foundation of China under Grant 61931022 and Grant 62271301; in part by the Shanghai Science and Technology Program under Grant 22511105200; in part by the Shanghai Excellent Academic Leaders Program under Grant 23XD1401400; in part by the Natural Science Foundation of Shandong Province under Grant ZR2022ZD38.

{\small
\bibliographystyle{unsrt}
\bibliography{reference}
}

\end{document}